\pdfoutput=1

\documentclass[11pt]{article}

\usepackage[final]{acl}

\usepackage{times}
\usepackage{latexsym}

\usepackage[T1]{fontenc}
\usepackage[utf8]{inputenc}
\usepackage{microtype}
\usepackage{inconsolata}
\usepackage{graphicx}

\usepackage{amsmath}
\usepackage{amsthm}
\usepackage{booktabs}

\numberwithin{lemma}{section}
\numberwithin{theorem}{section}

\numberwithin{corollary}{section}

\numberwithin{assumption}{section}
\usepackage{comment}
\usepackage{stfloats}
\usepackage{array}
\usepackage{multirow}
\usepackage{subcaption}
\usepackage{tcolorbox}
\tcbuselibrary{skins,breakable}
\usepackage{float}
\usepackage[table]{xcolor}

\usepackage{booktabs}
\usepackage{tabularx}
\usepackage{makecell}
\usepackage{ragged2e}
\usepackage{array}

\usepackage[table]{xcolor}
\usepackage{pifont}

\usepackage{pifont}

\usepackage{array}
\usepackage{multirow}
\usepackage{booktabs}
\usepackage[table]{xcolor}
\usepackage{makecell}
\usepackage{array}
\usepackage{amssymb}
\usepackage{amsfonts}
\usepackage{siunitx}

\newcolumntype{C}[1]{>{\centering\arraybackslash}m{#1}}
\newcolumntype{L}[1]{>{\raggedright\arraybackslash}m{#1}}

\title{FigEx2: Visual-Conditioned Panel Detection and Captioning for Scientific Compound Figures}

\author{ 
    Jifeng Song\textsuperscript{1, 2},
    Arun Das\textsuperscript{2, 3},
    Pan Wang\textsuperscript{1},
    Hui Ji\textsuperscript{4},
    Kun Zhao\textsuperscript{1},
    Yufei Huang\textsuperscript{1, 2, 3}\thanks{\,\small Corresponding author}\;\\ 
\textsuperscript{1}\,Department of Electrical and Computer Engineering, University of Pittsburgh, USA\vspace{-0.5mm} \\ 
\textsuperscript{2}\,Cancer Virology Program, UPMC Hillman Cancer Center, USA\vspace{-0.5mm} \\
\textsuperscript{3}\,Department of Medicine, University of Pittsburgh, USA\vspace{-0.5mm} \\
\textsuperscript{4}\,Department of Informatics and Networked Systems, University of Pittsburgh, USA\vspace{-0.5mm} \\
\texttt{
\{jis219, ard212, pan.wang, huj16, kun.zhao, yuh119\}@pitt.edu  
}
}

\begin{document}
\maketitle

\begin{abstract}
Scientific compound figures combine multiple labeled panels into a single image, and downstream pretraining and retrieval require panel-aligned visual-text pairs. However, in a PubMed Central (PMC)-scale crawl of 346,567 compound figures, 16.3\% have no caption and are discarded by existing caption-decomposition pipelines. We propose FigEx2, a visual-conditioned framework that takes only a compound figure as input and jointly produces labeled panel boxes and panel-wise captions. FigEx2 introduces an Entity-Attention Kullback--Leibler (KL) regularizer that aligns the detector's cross-attention with scientific entities annotated for each panel, providing a stable conditioning signal that also improves localization, and applies Group Relative Policy Optimization (GRPO) with a panel-level Entity-F1 reward to optimize scientific faithfulness. We curate BioSci-Fig-Cap for in-domain supervision and contribute physics and chemistry test suites for cross-disciplinary evaluation. FigEx2 achieves 0.751 mAP@0.5:0.95 on BioSci-Fig-Cap, and outperforms Qwen3-VL-8B by 6.80 Entity-F1 on MedICaT for captioning. It also transfers zero-shot to out-of-distribution domains. The source code is available at \url{https://github.com/Huang-AI4Medicine-Lab/FigEx2}.
\end{abstract}

\begin{figure}[t!]
  \centering
  \includegraphics[width=\linewidth]{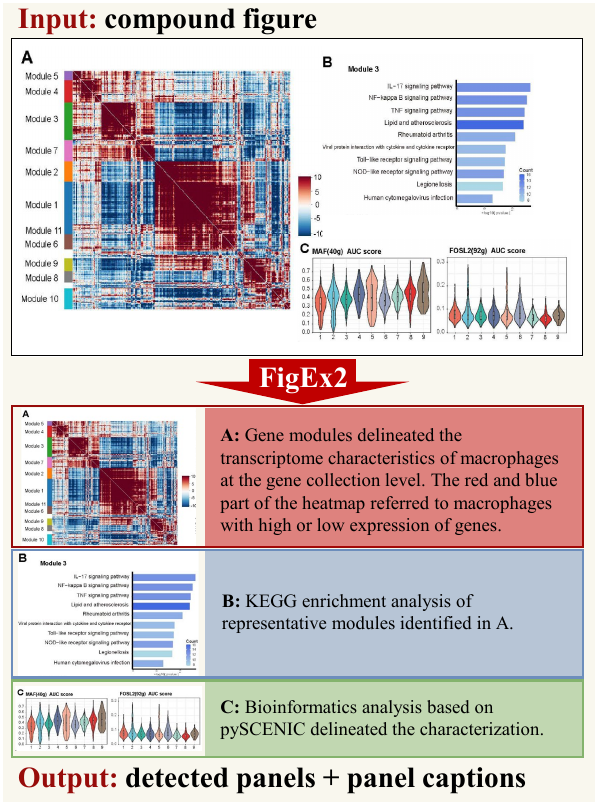}
  \caption{Task overview of FigEx2. Input is only the compound figure. FigEx2 detects labeled panels and generates panel-wise captions.}
  \label{fig:task_overview}
\end{figure}

\section{Introduction}

Scientific literature frequently employs compound figures \cite{lee2017viziometrics, subramanian2020medicat} that aggregate multiple panels into a single image. Since each panel often depicts a distinct experiment, condition, or analysis, fine-grained understanding requires both localizing these panels and generating panel-specific captions. In practice, however, panel-level captions are frequently missing at scale. Our goal is to produce panel-aligned visual-text pairs from compound figures even in the absence of caption supervision, increasing effective dataset coverage for downstream representation learning.

Previous work \cite{song2025figex} addressed panel extraction by formulating it as a caption separation task: aligning segments of a panel-resolved caption to predicted visual boxes. However, this relies on a high-fidelity text-to-image mapping that is not always available in real-world pipelines \cite{clark2016pdffigures}, and captions are often missing or limited to high-level summaries \cite{li2019figure} that fail to describe panels individually.

To resolve this, we propose a visual-conditioned formulation that does not assume caption availability. As illustrated in Figure~\ref{fig:task_overview}, the model takes only the compound figure as input, detects labeled panels, and generates a caption for each panel based on visual content, broadening the set of compound figures from which usable supervision can be extracted.

Based on this insight, we introduce FigEx2, a unified framework for visually grounded panel detection and captioning. Given only a compound figure, FigEx2 (i) localizes individual panels and (ii) generates panel-specific captions in a structured format that ends with a \texttt{[DET]} trigger \cite{wei2025lenna} to link captioning with detection. A central design question is how the generated caption should condition the detector. Key biological entities such as genes, cell types, and diseases typically appear as printed annotations inside the corresponding panel; compared to the descriptive language that fills most of a caption, the caption tokens that match these on-panel entities provide a much more localized signal for detector queries.

We operationalize this idea through an Entity-Attention Kullback--Leibler (KL) regularizer that pulls the detector's query-to-text cross-attention onto entity tokens during supervised training. After supervised fine-tuning, we further apply Group Relative Policy Optimization (GRPO) \cite{shao2024deepseekmath, liang2025group} with a panel-level Entity-F1 reward, directly optimizing the scientific content that prior generic similarity or embedding-based evaluation signals across generation tasks \cite{rennie2017self, zhang2019bertscore, hessel2021clipscore, zhao2023evaluating, zhao2024slide} fail to target.

To support high-quality supervision, we introduce BioSci-Fig-Cap, a curated panel-caption benchmark derived from BioSci-Fig \cite{song2025figex}, and contribute two out-of-domain test suites, PhysSci-Fig-Cap-Test (physics) and ChemSci-Fig-Cap-Test (chemistry), to benchmark cross-disciplinary transfer.

Our contributions are summarized as follows:
\begin{itemize}
    \item We reformulate panel extraction from caption decomposition to visual-conditioned panel detection and captioning, where FigEx2 takes only a compound figure and produces labeled panel captions linked with detection via a \texttt{[DET]} trigger.
    \item We propose an Entity-Attention KL that aligns the detector's query-to-text cross-attention with entity tokens in the generated caption, providing a stable conditioning signal for the caption-to-detection interface that improves panel detection in addition to captioning. This contrasts with prior entity-grounded captioning, which uses entities only as RL rewards.
    \item We optimize panel captioning with GRPO using a panel-level Entity-F1 reward that directly targets scientific faithfulness, and report Entity-F1 as a new evaluation dimension complementary to lexical and embedding-based metrics.
    \item We curate BioSci-Fig-Cap for cleaner panel-level supervision and build two out-of-domain test suites in physics and chemistry to benchmark cross-domain transfer.
\end{itemize}

\section{Related Work}

\subsection{Compound Figure Separation}
Decomposing a scientific figure that contains multiple panels into labeled panels is essential for panel-level understanding and evaluation. This task has evolved from hand-crafted separator rules to pixel-level deep learning (ImageCLEF~\cite{de2016overview}), with CNNs~\cite{tsutsui2017data} and label-guided pipelines~\cite{jiang2021two} significantly improving layout robustness. In the biomedical domain, SimCFS~\cite{yao2022compound} further leverages synthetic composition to mitigate high annotation costs. However, existing methods focus primarily on geometric partitioning, often decoupling spatial separation from downstream semantic reasoning. Our work bridges this gap by integrating separation into a unified framework that links panel detection directly to generative reasoning.

\subsection{Vision-Language Detection and Grounding}
Large-scale Vision-Language Models (VLMs) have recently achieved remarkable success in multimodal understanding, demonstrating impressive capabilities in generating and interpreting content across visual and textual modalities \cite{wang2026models}, with multimodal fusion and representation learning explored across tasks such as sentiment analysis \cite{wang2025dlf, wang2026consmsa}. Driven by these advancements, detection has evolved from contrastive region-text alignment (ViLD~\cite{gu2021open}, GLIP~\cite{li2022grounded}, DQ-DETR~\cite{liu2023dq}, Grounding DINO~\cite{liu2024grounding}) to generative grounding via coordinate tokenization (Pix2Seq~\cite{chen2021pix2seq}, OFA~\cite{wang2022ofa}). Modern MLLMs extend this through external routing (DetGPT~\cite{pi2023detgpt}, VisionLLM~\cite{wang2023visionllm}, LISA~\cite{lai2024lisa}) or native integration (Kosmos-2~\cite{peng2023kosmos}, CogVLM~\cite{wang2024cogvlm}, Lenna~\cite{wei2025lenna}). While foundation VLMs like Qwen2.5/3-VL~\cite{Qwen2.5-VL, Bai2025Qwen3VLTR} show robust spatial reasoning~\cite{li2025industrynav}, RL-optimized localization (VLRM~\cite{dzabraev2024vlrm}) remains nascent. Crucially, while these systems typically rely on textual priors, our work leverages DAB-DETR~\cite{liu2022dab} to enable high-fidelity multi-instance panel localization in scientific compound figures, where localization must stay consistent with a structured, label-conditioned captioning output.

\subsection{Entity-Grounded Caption Generation}
Closely related to our setting, recent work in radiology report generation uses entity-level rewards to improve factual correctness. Building on the RadGraph dataset of clinical entities and relations~\cite{jain2021radgraph}, \citet{delbrouck2022improving} introduce a RadGraph-based F1 reward that scores generated reports against reference reports on the set of clinical entities, and subsequent work extends this direction with semantic-graph combinations and multi-agent decomposition. These methods demonstrate that entity-level rewards substantially improve factual consistency over lexical or generic semantic rewards. Beyond reward design, radiology report generation has also been paired with more robust evaluation protocols~\cite{zhao2026x}, underscoring that faithful generation and reliable evaluation are tightly coupled.

However, these methods use entities only at the reward level, a sequence-level signal applied during reinforcement learning, leaving the underlying representations unconstrained with respect to entity content. They also confine entity supervision to a single domain (chest X-ray), and only target caption faithfulness without considering how entity information could support spatial localization within the image.

\section{Method}
\label{sec:method}

\begin{figure*}[t]
  \centering
  \includegraphics[width=\linewidth]{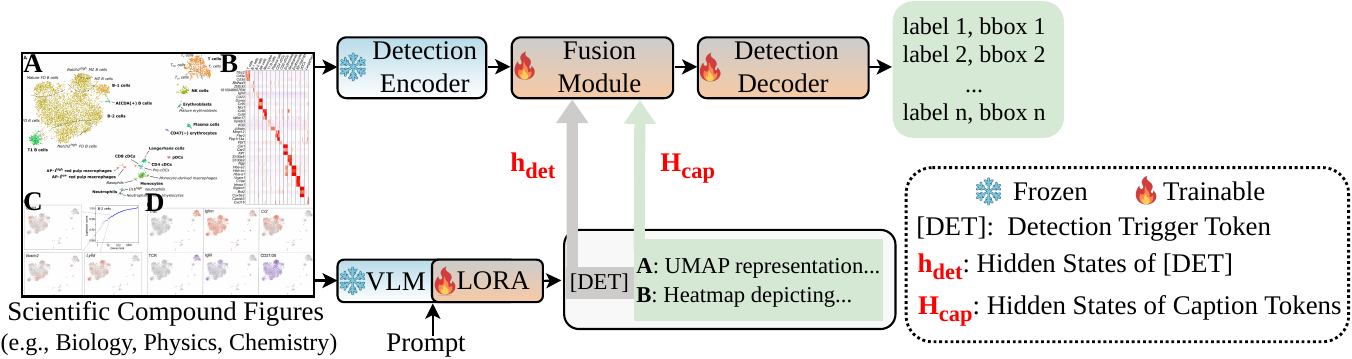}
  \caption{Overall framework of FigEx2. Given a compound figure, the captioning branch generates panel-wise captions ending with a \texttt{[DET]} token. The hidden state at \texttt{[DET]} ($\mathbf{h}_{det}$) and the hidden states of the remaining generated caption tokens ($\mathbf{H}_{cap}$) are passed to the Fusion Module, which integrates them with visual features from the detection encoder; the detection decoder then predicts panel labels and bounding boxes.}
  \label{fig:framework}
\end{figure*}

\subsection{Overview}
Given an input compound figure and a prompt, the captioning branch (a frozen vision-language model with trainable LoRA adapters) generates panel-wise captions terminated by a [DET] token. The detection encoder independently encodes the same compound figure. Two caption-side features are then passed into the Fusion Module: the hidden state at [DET] ($\mathbf{h}_{det}$), and the hidden states of the remaining generated caption tokens ($\mathbf{H}_{cap}$). The Fusion Module integrates these with the visual features from the detection encoder, and the detection decoder predicts panel labels and bounding boxes.

\subsection{Captioning to Detection Interface}
\label{sec:cap_det_interface}
Let the captioning branch produce a token sequence $\hat{y}_{1:T}$ that includes a special trigger token \texttt{[DET]}, with final-layer hidden states $\mathbf{H}\in\mathbb{R}^{T\times d}$, where $T$ is the output length and $d$ is the hidden size. We denote the hidden state at the \texttt{[DET]} position as $\mathbf{h}_{det}\in\mathbb{R}^{d}$.

We expose the detector to the full sequence of generated caption tokens (excluding \texttt{[DET]}). Let $\mathbf{H}_{cap}\in\mathbb{R}^{T'\times d}$ denote the stacked hidden states of these $T'$ tokens. We project these features into the detector's text space:
\begin{equation}
\mathbf{F}_{txt} = \mathbf{H}_{cap}\mathbf{W}_{txt} \in \mathbb{R}^{T'\times d},
\label{eq:ftxt_def}
\end{equation}
where $\mathbf{W}_{txt}\in\mathbb{R}^{d\times d}$ is a learned linear projection. Together with the \texttt{[DET]} hidden state $\mathbf{h}_{det}$, $\mathbf{F}_{txt}$ forms the caption-side input to the fusion module described next.

\subsection{Fusion Module}
\label{sec:fusion_module}
The Fusion Module integrates the caption-side features $\mathbf{h}_{det}$ and $\mathbf{F}_{txt}$ into the detector's object queries. Let $\mathbf{Q}\in\mathbb{R}^{N_d\times d}$ be the $N_d$ object queries initialized by the detection encoder, and let $\mathbf{F}_{det}=\mathbf{h}_{det}^{\top}\in\mathbb{R}^{1\times d}$ denote the detection-side interface token. We update the queries with two cross-attention steps:
\begin{align}
\mathbf{Q}' &= \textsc{Attn}(\mathbf{Q},\, \mathbf{F}_{det}), \label{eq:det_guided_attn}\\
\mathbf{Q}''&= \textsc{Attn}(\mathbf{Q}',\, \mathbf{F}_{txt}), \label{eq:text_guided_attn}
\end{align}
where $\textsc{Attn}(\mathbf{Q},\mathbf{F})$ is a standard multi-head attention block using $\mathbf{Q}$ as queries and $\mathbf{F}$ as keys/values. The detection decoder then consumes $\mathbf{Q}''$ and predicts a set of panel bounding boxes with standard query-based decoding.

\begin{figure}[t]
  \centering
  \includegraphics[width=\linewidth]{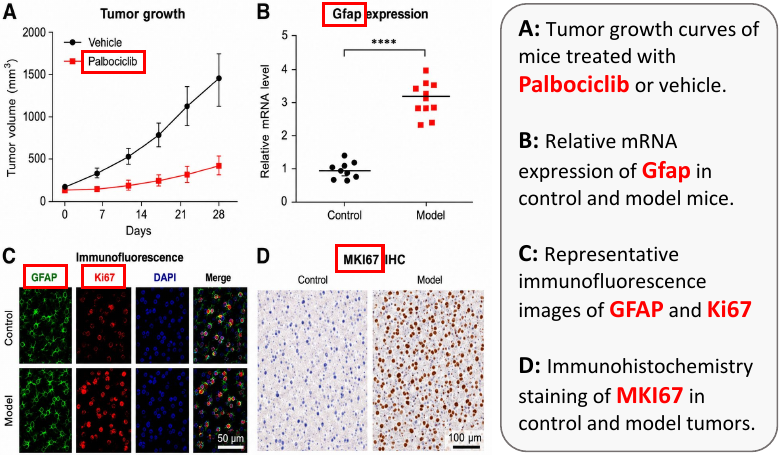}
  \caption{Scientific entities mentioned in panel captions (red) often appear as printed annotations inside the corresponding panel.}
  \label{fig:entity_anchor}
\end{figure}

\subsection{Entity-Attention Regularization}
\label{sec:entity_attn_kl}
Feeding the detector caption-token hidden states gives the model maximum information, but provides no inductive bias for which tokens carry panel-specific scientific content. As illustrated in Figure~\ref{fig:entity_anchor}, scientific entities often correspond to printed annotations inside the corresponding panel, making them natural visual anchors for detector queries. We therefore regularize the detector's cross-attention to concentrate on entity tokens.

\begin{figure}[t]
  \centering
  \includegraphics[width=\linewidth]{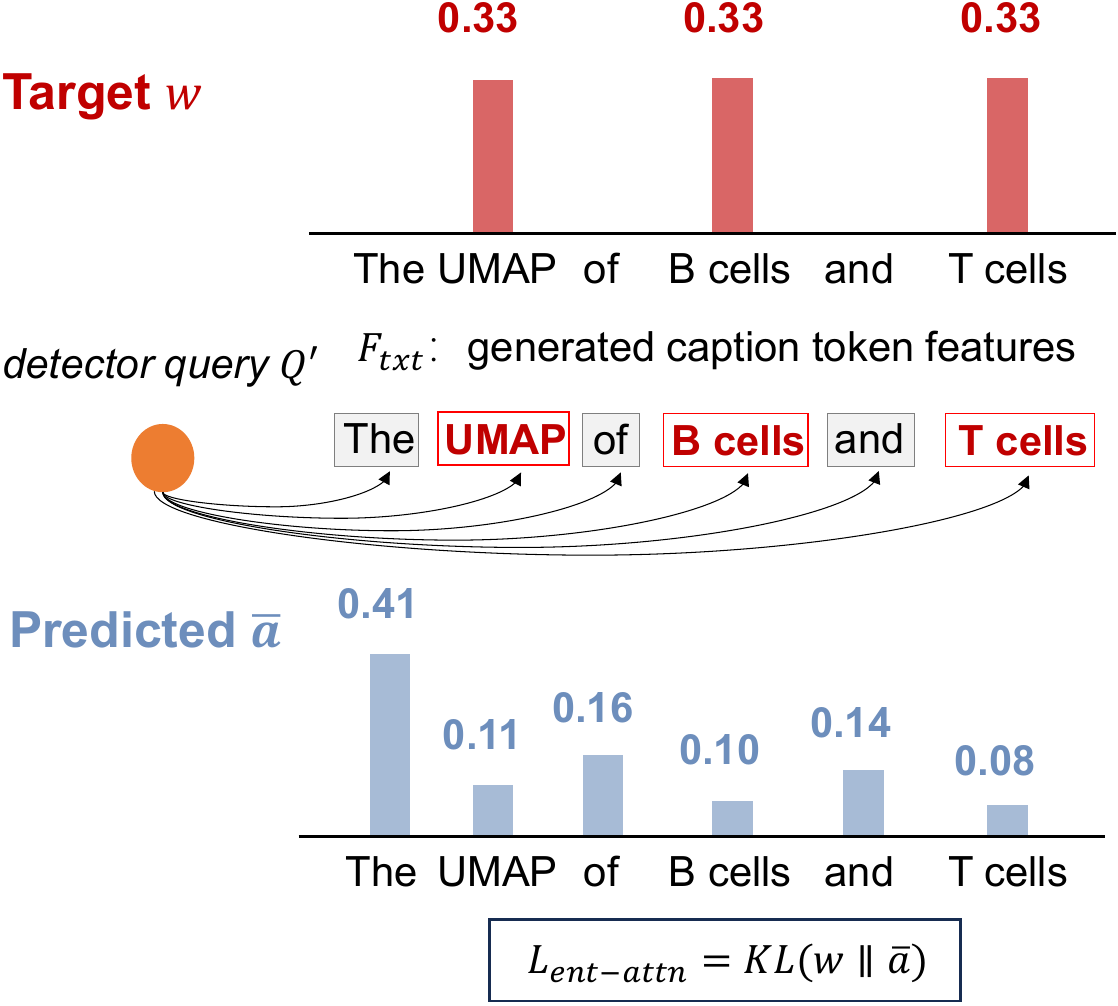}
  \caption{Entity-Attention KL on an example caption. The target $w$ (top) places uniform mass on entity tokens and zero mass on non-entity tokens. The KL pulls the predicted attention $\bar{a}$ (bottom) toward $w$.}
  \label{fig:entity_attn_kl}
\end{figure}

Let $\boldsymbol{\alpha}\in\mathbb{R}^{N_d\times T'}$ denote the softmax-normalized attention weights produced by Eq.~\eqref{eq:text_guided_attn}, where $\alpha_{i,t}$ is the attention from object query $i$ to caption token $t$. To exploit this property, we introduce a binary Entity-Attention KL that explicitly directs cross-attention mass toward entity tokens. Let $\bar{a}_t = \frac{1}{N_d}\sum_{i=1}^{N_d}\alpha_{i,t}$ denote the average attention from all object queries to caption token $t$, and define a target distribution $w$ supported only on the entity subset $\mathcal{T}_{ent}$:
\begin{equation}
w_t =
\begin{cases}
1/N_e & t\in\mathcal{T}_{ent}, \\
0 & t\notin\mathcal{T}_{ent}.
\end{cases}
\label{eq:kl_target}
\end{equation}
That is, non-entity caption tokens receive zero target mass, while the $N_e$ entity tokens share the mass uniformly. The regularizer is the KL divergence from $w$ to the predicted attention distribution:
\begin{equation}
\mathcal{L}_{\text{ent-attn}} = \mathrm{KL}\bigl(w \,\|\, \bar{a}\bigr) = \sum_{t\in\mathcal{T}_{ent}} w_t\log\frac{w_t}{\bar{a}_t + \varepsilon},
\label{eq:kl_loss}
\end{equation}
where $\varepsilon$ is a small constant for numerical stability. Figure~\ref{fig:entity_attn_kl} illustrates the construction on an example caption.

\subsection{Training Procedure}
\label{sec:training_procedure}

FigEx2 is trained to (i) generate structured panel captions and (ii) localize labeled panels. Given a compound figure $x$, the captioning branch generates a label-ordered target sequence $y_{1:T}$ that ends with the trigger token \texttt{[DET]}; conditioned on the caption hidden states, the detector predicts a set of panel boxes and labels. We use a four-stage training schedule.

\paragraph{Stage~1: Captioning pretraining.}
We freeze the detector and update only the captioning branch with token-level cross entropy:
\begin{equation}
\mathcal{L}_{cap} = -\frac{1}{T}\sum_{t=1}^{T}\log p\bigl(y_t \mid y_{<t}, x\bigr),
\label{eq:cap_ce}
\end{equation}
where $y_{1:T}$ is the structured ground-truth caption sequence of 
length $T$.

\paragraph{Stage~2: Detection pretraining.}
We freeze the captioning branch and train the detector with a standard set-based detection objective:
\begin{equation}
\mathcal{L}_{det}
= \lambda_{cls}\mathcal{L}_{cls}
+ \lambda_{bbox}\mathcal{L}_{bbox}
+ \lambda_{giou}\mathcal{L}_{giou},
\label{eq:det_loss}
\end{equation}
where $\mathcal{L}_{cls}$ is the classification loss, $\mathcal{L}_{bbox}$ is the box regression loss, $\mathcal{L}_{giou}$ is the generalized IoU loss, and $\lambda_{cls}$, $\lambda_{bbox}$, and $\lambda_{giou}$ are scalar weights.

\paragraph{Stage~3: Joint supervised training with Entity-Attention KL.}
Starting from the Stage~1 and Stage~2 checkpoints, we jointly optimize 
captioning, detection, and the Entity-Attention KL:
\begin{equation}
\mathcal{L}_{sft} = \lambda_{cap}\mathcal{L}_{cap} + \lambda_{det}\mathcal{L}_{det} 
+ \lambda_{\text{ent-attn}}\mathcal{L}_{\text{ent-attn}},
\label{eq:sft_total}
\end{equation}
where $\mathcal{L}_{cap}$, $\mathcal{L}_{det}$, and 
$\mathcal{L}_{\text{ent-attn}}$ are the losses defined in 
Eqs.~(\ref{eq:cap_ce}), (\ref{eq:det_loss}), and (\ref{eq:kl_loss}), 
and $\lambda_{cap}$, $\lambda_{det}$, $\lambda_{\text{ent-attn}}$ are 
scalar loss weights.

\paragraph{Stage~4: Reward-augmented training with GRPO.}
\label{para:stage4}
Starting from the Stage~3 checkpoint, we apply Group Relative Policy 
Optimization (GRPO)~\cite{shao2024deepseekmath} to the captioning branch 
with a LoRA adapter, while the detector remains frozen. For each 
input, we sample $K$ candidates $\{\hat{Y}_i\}_{i=1}^{K}$ from the 
behavior policy $\pi_{\theta_{old}}$ (the captioning policy at the start 
of each GRPO update), compute their sequence-level rewards, and form a 
group-normalized advantage $A_i$ by standardizing rewards within the 
group. The objective is a PPO-style clipped policy gradient with a KL 
penalty against the Stage~3 reference $\pi_{ref}$:
\begin{equation}
\begin{aligned}
\mathcal{L}_{rl} = & -\frac{1}{K}\sum_{i=1}^{K}\frac{1}{|\hat{Y}_i|}\sum_{t=1}^{|\hat{Y}_i|}\min\bigl(\rho_{i,t}A_i,\\
&\mathrm{clip}(\rho_{i,t},1-\epsilon,1+\epsilon)A_i\bigr) + \beta\,\mathrm{KL}(\pi_\theta\,\|\,\pi_{ref}),
\end{aligned}
\label{eq:grpo_obj}
\end{equation}
where $\hat{Y}_i$ is the $i$-th sampled caption with token length 
$|\hat{Y}_i|$, $\rho_{i,t}=\pi_\theta(\hat{y}_{i,t}\mid x,\hat{y}_{i,<t})/\pi_{\theta_{old}}(\hat{y}_{i,t}\mid x,\hat{y}_{i,<t})$ 
is the per-token importance ratio between the current and behavior 
policies, $\epsilon$ is the PPO clip range, and $\beta$ controls the 
strength of the KL penalty.

\paragraph{Entity-F1 reward.}
The reward is panel-level Entity-F1. Each sampled caption is parsed into label-keyed subcaptions, and for each ground-truth panel label we run the dataset-specific entity extractor (Appendix~\ref{app:entity_extractors}) on the predicted and reference subcaptions and compute set-level F1 between the resulting entity sets. 

\section{Benchmarks and Curation}\label{sec:datasets_cap}
We evaluate on four benchmarks. \textbf{BioSci-Fig-Cap} is a curated panel-caption benchmark we derive from BioSci-Fig with cleaner panel-level supervision; it is used for in-domain training and evaluation. \textbf{PhysSci-Fig-Cap-Test} (physics) and \textbf{ChemSci-Fig-Cap-Test} (chemistry) are two test-only benchmarks we curate to evaluate cross-disciplinary transfer. We also report results on \textbf{MedICaT}~\cite{subramanian2020medicat} for comparison with prior compound-figure protocols. BioSci-Fig-Cap is curated within each original BioSci-Fig subset, with its test set sourced only from the original test subset. Table~\ref{tab:dataset_split_cap} reports dataset sizes. Curation details and per-dataset entity extractors are provided in Appendices~\ref{app:benchmark_details} and~\ref{app:entity_extractors}.

\begin{table}[t]
\centering
\small
\setlength{\tabcolsep}{5.5pt}
\renewcommand{\arraystretch}{1.0}
\begin{tabular*}{\columnwidth}{@{\extracolsep{\fill}} l l S[table-format=4.0] S[table-format=5.0]}
\toprule
\textbf{Dataset} & \textbf{Split} & \textbf{Figures} & \textbf{Pairs} \\
\midrule
\multirow{3}{*}{MedICaT} & Train & 1671 & 6359 \\
                         & Val   & 209  & 776  \\
                         & Test  & 209  & 840  \\
\midrule[0.4pt]
\multirow{3}{*}{BioSci-Fig-Cap} & Train & 5290 & 30865 \\
                               & Val   & 652  & 3783  \\
                               & Test  & 658  & 3959  \\
\midrule[0.4pt]
PhysSci-Fig-Cap-Test & Test & 200 & 778 \\
\midrule[0.4pt]
ChemSci-Fig-Cap-Test & Test & 200 & 793 \\
\bottomrule
\end{tabular*}
\caption{Dataset splits after curation. We report the number of compound figures and panel-caption pairs.}
\label{tab:dataset_split_cap}
\end{table}

\section{Experiments}
\label{sec:experiments}

\subsection{Evaluation Metrics}\label{sec:metrics}
We report mAP@0.5:0.95 and mAP@0.5 for panel detection. For panel captioning, we use BLEU-4, ROUGE-L, METEOR, BERTScore, and Entity-F1. Entity-F1 is computed in the same way as the Stage~4 reward (Section~\ref{para:stage4}), but is applied at evaluation time to compare predicted and reference panel captions. Predictions are aligned with ground-truth panels by label in reading order, with missing or extra pairs scored as 0; full details are in Appendix~\ref{app:eval_protocol}. 

\subsection{Baselines}\label{sec:baselines}
We compare against three baseline families: vision-only detectors~\cite{fang2021you} for panel detection, vision-language captioners~\cite{liu2024llavanext, li2023llava, grattafiori2024llama} prompted to produce labeled panel captions, and vision-language box generators prompted to output structured bounding boxes that we parse and evaluate as detections. Prompts are provided in Appendix~\ref{app:prompts}.

\subsection{Implementation Details}\label{sec:impl}
FigEx2 combines Qwen3-VL-8B with a DAB-DETR detector and is trained with the four-stage procedure in Section~\ref{sec:training_procedure}. We train on 4 NVIDIA A100 80GB GPUs and run inference on NVIDIA L40S 48GB and NVIDIA A100 40GB GPUs. The hyperparameter settings are in Appendix~\ref{app:impl}.

\subsection{Comparison with Existing Methods}
\label{sec:comparison}

\begin{table*}[t]
\centering
\scriptsize
\setlength{\tabcolsep}{3pt}
\renewcommand{\arraystretch}{1.05}
\resizebox{\textwidth}{!}{%
\begin{tabular}{l c c c c c c c c c c}
\toprule
\textbf{Model} &
\multicolumn{5}{c}{\textbf{MedICaT}} &
\multicolumn{5}{c}{\textbf{BioSci-Fig-Cap}} \\
\cmidrule(lr){2-6}\cmidrule(lr){7-11}
& \textbf{BLEU-4} & \textbf{ROUGE-L} & \textbf{METEOR} & \textbf{BERTScore} & \textbf{Entity-F1}
& \textbf{BLEU-4} & \textbf{ROUGE-L} & \textbf{METEOR} & \textbf{BERTScore} & \textbf{Entity-F1} \\
\midrule
\rowcolor{gray!15}
\multicolumn{11}{l}{\textbf{Pre-trained}}\\
Qwen2.5-VL-72B & 2.08 & 15.91 & 9.95 & 83.67 & 9.95 & 1.71 & 23.62 & 14.74 & 82.23 & 21.36 \\
Llama3.2-90B   & 3.00 & 11.73 & 7.97 & 85.68 & 9.40 & 2.94 & 18.35 & 11.80 & 85.96 & 24.39 \\
gpt-5-nano     & 2.01 & 10.76 & 5.91 & 85.23 & 5.65 & 2.84 & 16.97 & 9.76  & 85.79 & 27.57 \\
gpt-5-mini     & 2.92 & 14.44 & 9.38 & 85.44 & 9.52 & 4.44 & 23.03 & 16.20 & 86.67 & 35.10 \\
gpt-5.2        & 2.43 & 13.25 & 7.44 & 85.54 & 7.15 & 4.07 & 23.85 & 14.63 & 87.12 & 35.56 \\
\addlinespace[2pt]
\rowcolor{gray!15}
\multicolumn{11}{l}{\textbf{Fine-tuned}}\\
LLaVA-NeXT     & 4.06 & 14.27 & 9.28  & 85.14 & 9.12  & 3.36 & 20.20 & 12.74 & 86.22 & 17.67 \\
LLaVA-Med      & 2.26 & 10.58 & 6.26  & 85.00 & 6.05  & 2.91 & 17.57 & 11.32 & 85.78 & 13.44 \\
Llama3.2-11B   & 4.80 & 16.38 & 11.39 & 85.62 & 8.43  & 3.88 & 21.44 & 15.79 & 86.33 & 22.58 \\
Qwen3-VL-8B    & 7.37 & 18.13 & 12.78 & 86.25 & 6.58  & 5.21 & 25.69 & 18.27 & 87.00 & 27.09 \\
FigEx2-8B      & \textbf{9.05} & \textbf{22.58} & \textbf{15.81} & \textbf{86.53} & \textbf{13.38}
               & \textbf{5.56} & \textbf{26.63} & \textbf{19.05} & \textbf{87.38} & \textbf{36.36} \\
\bottomrule
\end{tabular}%
}
\caption{Panel captioning on MedICaT and BioSci-Fig-Cap.}
\label{tab:captioning_main_merged}
\end{table*}

\textbf{Panel Captioning.}
Table~\ref{tab:captioning_main_merged} reports panel captioning results. FigEx2 achieves the best overall performance on both datasets across all five metrics. On MedICaT, FigEx2 improves ROUGE-L from 18.13 to 22.58 and METEOR from 12.78 to 15.81 over Qwen3-VL-8B, the strongest fine-tuned baseline, while raising Entity-F1 from 6.58 to 13.38, more than doubling the Entity-F1 of Qwen3-VL-8B. On BioSci-Fig-Cap, FigEx2 again improves all five metrics over Qwen3-VL-8B, with the most pronounced gain on Entity-F1 (27.09 to 36.36). The disproportionately large gains on Entity-F1 indicate that FigEx2's predictions recover scientific entities much more reliably than baseline captions of comparable lexical quality.

\textbf{Panel Detection.}
Table~\ref{tab:panel_detection_singlecol} reports panel detection results under the LVIS evaluator~\cite{gupta2019lvis, tran2025simltd}. On MedICaT, FigEx2 achieves the highest mAP, improving mAP@0.5:0.95 from 0.230 to 0.360 over prompt-based box generation with Qwen3-VL-8B, and mAP@0.5 from 0.309 to 0.405. On BioSci-Fig-Cap, FigEx2 remains the top method, improving mAP@0.5:0.95 from 0.697 to 0.751 over DAB-DETR and mAP@0.5 from 0.830 to 0.906. The detection gains are notable given that DAB-DETR is already a strong specialized detector on BioSci-Fig-Cap (0.697 mAP@0.5:0.95); coupling the same detector backbone with caption-conditioned object queries lifts panel localization across both datasets.

\label{sec:comparison_det}

\begin{table}[t]
\centering
\small
\renewcommand{\arraystretch}{1.10}
\setlength{\tabcolsep}{5pt}
\begin{tabular}{lcc}
\toprule
\textbf{Model} & \textbf{mAP@0.5:0.95} & \textbf{mAP@0.5} \\
\midrule

\rowcolor{gray!15}\multicolumn{3}{l}{\textbf{MedICaT}}\\
YOLOS-Ti       & 0.155 & 0.183 \\
YOLOS-S        & 0.156 & 0.188 \\
YOLOS-B        & 0.150 & 0.170 \\
DAB-DETR       & 0.165 & 0.181 \\
Qwen3-VL-8B     & 0.230 & 0.309 \\
FigEx2-8B      & \textbf{0.360} & \textbf{0.405} \\
\addlinespace[2pt]

\rowcolor{gray!15}\multicolumn{3}{l}{\textbf{BioSci-Fig-Cap}}\\
YOLOS-Ti       & 0.100 & 0.129 \\
YOLOS-S        & 0.196 & 0.248 \\
YOLOS-B        & 0.448 & 0.534 \\
DAB-DETR       & 0.697 & 0.830 \\
Qwen3-VL-8B     & 0.439 & 0.578 \\
FigEx2-8B      & \textbf{0.751} & \textbf{0.906} \\
\bottomrule
\end{tabular}
\caption{Panel detection on MedICaT and BioSci-Fig-Cap.}
\label{tab:panel_detection_singlecol}
\end{table}

\subsection{Ablation Study}

\begin{table*}[!t]
\centering
\scriptsize
\renewcommand{\arraystretch}{1.05}
\setlength{\tabcolsep}{3.5pt}
\resizebox{\textwidth}{!}{%
\begin{tabular}{@{} l cc ccccc @{}}
\toprule
\textbf{Variant}
& \multicolumn{2}{c}{\textbf{Panel Detection}}
& \multicolumn{5}{c}{\textbf{Panel Captioning}} \\
\cmidrule(lr){2-3}\cmidrule(lr){4-8}
& \textbf{mAP@0.5:0.95} & \textbf{mAP@0.5}
& \textbf{BLEU-4} & \textbf{ROUGE-L} & \textbf{METEOR} & \textbf{BERTScore} & \textbf{Entity-F1} \\
\midrule
\rowcolor{gray!15}\multicolumn{8}{l}{\textbf{MedICaT}}\\
(1) SFT only, no Entity-Attention KL              & 0.272 & 0.301 & 8.11 & 21.17 & 14.58 & 86.45 & 7.45 \\
(2) SFT + Entity-Attention KL                     & 0.353 & 0.403 & 8.54 & 21.63 & 15.44 & \textbf{86.54} & 9.88 \\
(3) SFT + Entity-Attention KL + GRPO (Entity-F1)  & \textbf{0.360} & \textbf{0.405} & \textbf{9.05} & \textbf{22.58} & \textbf{15.81} & 86.53 & \textbf{13.38} \\
\addlinespace[1pt]
\rowcolor{gray!15}\multicolumn{8}{l}{\textbf{BioSci-Fig-Cap}}\\
(1) SFT only, no Entity-Attention KL              & 0.722 & 0.865 & 5.35 & 25.75 & 18.25 & 87.15 & 31.85 \\
(2) SFT + Entity-Attention KL                     & 0.740 & 0.881 & 5.45 & 26.16 & 18.89 & 87.27 & 34.63 \\
(3) SFT + Entity-Attention KL + GRPO (Entity-F1)  & \textbf{0.751} & \textbf{0.906} & \textbf{5.56} & \textbf{26.63} & \textbf{19.05} & \textbf{87.38} & \textbf{36.36} \\
\bottomrule
\end{tabular}%
}
\caption{Ablation of FigEx2 on MedICaT and BioSci-Fig-Cap. (1) SFT-only baseline; (2) adds the Entity-Attention KL; (3) further adds GRPO with the panel-level Entity-F1 reward.}
\label{tab:ablation_unified}
\end{table*}

Table~\ref{tab:ablation_unified} reports our ablation across two design choices: the Entity-Attention KL on detector cross-attention, and the GRPO stage. Adding the Entity-Attention KL improves panel detection mAP@0.5:0.95 from 0.272 to 0.353 on MedICaT and from 0.722 to 0.740 on BioSci-Fig-Cap, and improves panel captioning across all metrics, with Entity-F1 rising from 7.45 to 9.88 on MedICaT and from 31.85 to 34.63 on BioSci-Fig-Cap. Notably the KL also benefits captioning even though it is applied only on the detector side: gradients flow back through the shared caption features, signaling to the captioning branch that entity tokens carry weight. Adding GRPO with the Entity-F1 reward further raises Entity-F1 to 13.38 on MedICaT and 36.36 on BioSci-Fig-Cap, with consistent gains on BLEU-4, ROUGE-L, and METEOR; detection improves as well, reaching 0.751 mAP@0.5:0.95 on BioSci-Fig-Cap, since richer entity content in the generated captions feeds cleaner anchors back to the detector. The two components therefore form a positive loop: the KL makes the detector rely on entity tokens, which pushes captioning to produce them more reliably, and GRPO directly reinforces this behavior. Captioning and detection co-improve rather than trade off.

\begin{table*}[t]
\centering
\scriptsize
\renewcommand{\arraystretch}{1.05}
\setlength{\tabcolsep}{2pt}
\resizebox{\textwidth}{!}{%
\begin{tabular}{l c c c c c c c c c c}
\toprule
\textbf{Model} &
\multicolumn{5}{c}{\textbf{PhysSci-Fig-Cap-Test}} &
\multicolumn{5}{c}{\textbf{ChemSci-Fig-Cap-Test}} \\
\cmidrule(lr){2-6}\cmidrule(lr){7-11}
& \textbf{BLEU-4} & \textbf{ROUGE-L} & \textbf{METEOR} & \textbf{BERTScore} & \textbf{Entity-F1}
& \textbf{BLEU-4} & \textbf{ROUGE-L} & \textbf{METEOR} & \textbf{BERTScore} & \textbf{Entity-F1} \\
\midrule
\rowcolor{gray!15}
\multicolumn{11}{l}{\textbf{Pre-trained}}\\
Qwen2.5-VL-72B & 1.48 & 17.28 & 9.97  & 82.11 & 9.29  & 1.38 & 17.68 & 10.37 & 81.64 & 8.38  \\
Llama3.2-90B   & 2.09 & 12.84 & 8.05  & 84.76 & 11.85 & 2.21 & 14.61 & 8.99  & 84.45 & 12.98 \\
gpt-5-nano     & 1.85 & 11.56 & 6.33  & 84.37 & 9.75  & 1.73 & 11.60 & 6.09  & 84.15 & 13.00 \\
gpt-5-mini     & \textbf{2.81} & 16.11 & \textbf{10.56} & 84.81 & 14.55 & 2.82 & 16.41 & 10.44 & 84.47 & 14.34 \\
gpt-5.2        & 2.77 & 16.71 & 9.75  & 85.34 & 15.55 & 2.53 & 17.16 & 9.57  & 85.12 & 15.03 \\
\addlinespace[2pt]
\rowcolor{gray!15}
\multicolumn{11}{l}{\textbf{Fine-tuned}}\\
LLaVA-NeXT     & 1.48 & 11.25 & 6.07  & 84.45 & 6.03  & 1.93 & 14.54 & 8.56  & 84.50 & 5.16  \\
LLaVA-Med      & 1.56 & 10.72 & 6.04  & 84.67 & 1.64  & 1.57 & 12.83 & 7.75  & 84.43 & 4.44  \\
Llama3.2-11B   & 2.17 & 15.12 & 9.09  & 85.15 & 8.93  & 2.37 & 16.56 & 10.29 & 84.84 & 13.10 \\
Qwen3-VL-8B    & 2.75 & \textbf{17.75} & 9.98  & \textbf{85.57} & 12.28 & 3.05 & 18.93 & 11.58 & 85.18 & 14.78 \\
FigEx2-8B      & 2.65 & 16.90 & 9.39 & 85.50 & \textbf{16.38} & \textbf{3.05} & \textbf{19.92} & \textbf{12.08} & \textbf{85.39} & \textbf{15.29} \\
\bottomrule
\end{tabular}%
}
\caption{Cross-domain panel captioning on PhysSci-Fig-Cap-Test and ChemSci-Fig-Cap-Test.}
\label{tab:crossdomain_cap}
\end{table*}

\begin{table}[t]
\centering
\small
\renewcommand{\arraystretch}{1.05}
\setlength{\tabcolsep}{5pt}
\begin{tabular}{lcc}
\toprule
\textbf{Model} & \textbf{mAP@0.5:0.95} & \textbf{mAP@0.5} \\
\midrule

\rowcolor{gray!15}\multicolumn{3}{l}{\textbf{PhysSci-Fig-Cap-Test}}\\
YOLOS-Ti    & 0.087 & 0.107 \\
YOLOS-S     & 0.136 & 0.164 \\
YOLOS-B     & 0.237 & 0.280 \\
DAB-DETR    & 0.372 & 0.418 \\
Qwen3-VL-8B & 0.374 & 0.447 \\
FigEx2-8B   & \textbf{0.453} & \textbf{0.515} \\
\addlinespace[2pt]

\rowcolor{gray!15}\multicolumn{3}{l}{\textbf{ChemSci-Fig-Cap-Test}}\\
YOLOS-Ti    & 0.109 & 0.144 \\
YOLOS-S     & 0.185 & 0.231 \\
YOLOS-B     & 0.306 & 0.376 \\
DAB-DETR    & 0.357 & 0.411 \\
Qwen3-VL-8B & 0.322 & 0.431 \\
FigEx2-8B   & \textbf{0.446} & \textbf{0.511} \\
\bottomrule
\end{tabular}
\caption{Cross-domain panel detection on PhysSci-Fig-Cap-Test and ChemSci-Fig-Cap-Test.}
\label{tab:crossdomain_det}
\end{table}

\subsection{Cross-domain Generalization}
\label{sec:crossdomain}
We evaluate robustness under domain shift by transferring models fine-tuned on BioSci-Fig-Cap to PhysSci-Fig-Cap-Test and ChemSci-Fig-Cap-Test with no additional training. All detection results are obtained with detectors fine-tuned on BioSci-Fig-Cap, while captioning compares (i) pre-trained vision-language models and (ii) models fine-tuned on BioSci-Fig-Cap.

\begin{table*}[t]
\centering
\scriptsize
\renewcommand{\arraystretch}{1.05}
\setlength{\tabcolsep}{2.5pt}
\resizebox{\textwidth}{!}{%
\begin{tabular}{>{\centering\arraybackslash}p{1.4cm} ccccc ccccc}
\toprule
\textbf{Few-shot} &
\multicolumn{5}{c}{\textbf{PhysSci-Fig-Cap-Test}} &
\multicolumn{5}{c}{\textbf{ChemSci-Fig-Cap-Test}} \\
\cmidrule(lr){2-6}\cmidrule(lr){7-11}
&
\textbf{BLEU-4} & \textbf{ROUGE-L} & \textbf{METEOR} & \textbf{BERTScore} & \textbf{Entity-F1} &
\textbf{BLEU-4} & \textbf{ROUGE-L} & \textbf{METEOR} & \textbf{BERTScore} & \textbf{Entity-F1} \\
\midrule
\rowcolor{gray!15}\multicolumn{11}{l}{\textbf{Qwen3-VL-8B}}\\
1-shot & 20.10 & 33.48 & 28.60 & 89.57 & 16.73 & 32.63 & 46.17 & 42.46 & 91.24 & 28.23 \\
2-shot & \textbf{39.84} & \textbf{53.19} & \textbf{49.90} & 92.06 & 31.76 & 42.56 & 55.57 & 53.05 & 92.79 & 40.95 \\
\addlinespace[2pt]
\rowcolor{gray!15}\multicolumn{11}{l}{\textbf{FigEx2-8B}}\\
1-shot & 19.97 & 32.83 & 27.94 & 89.76 & 19.72 & 32.48 & 45.58 & 41.91 & 91.68 & 27.12 \\
2-shot & 38.36 & 50.99 & 48.06 & \textbf{92.22} & \textbf{35.48}
       & \textbf{46.90} & \textbf{59.96} & \textbf{56.90} & \textbf{93.39} & \textbf{43.81} \\
\bottomrule
\end{tabular}%
}
\caption{Few-shot prompting for cross-domain panel captioning on PhysSci-Fig-Cap-Test \& ChemSci-Fig-Cap-Test.}
\label{tab:fewshot_phy_chem}
\end{table*}

\paragraph{Cross-domain panel captioning.}
Table~\ref{tab:crossdomain_cap} summarizes cross-domain panel captioning results. On ChemSci-Fig-Cap-Test, which shares substantial domain knowledge with the biomedical training domain, FigEx2 achieves the best or tied-best result across all five metrics, improving ROUGE-L from 18.93 to 19.92, METEOR from 11.58 to 12.08, and Entity-F1 from 14.78 to 15.29 over Qwen3-VL-8B. On the more distant PhysSci-Fig-Cap-Test, FigEx2 trades a small amount of lexical overlap for a substantial gain in Entity-F1 from 12.28 to 16.38. The contrast suggests that domain proximity benefits lexical generation, but the Entity-Attention KL keeps the model focused on scientifically meaningful tokens even under stronger domain shift, so the entity-level advantage persists across both domains.

\paragraph{Cross-domain panel detection.}
Table~\ref{tab:crossdomain_det} reports cross-domain detection performance. Among classical detectors, DAB-DETR is the strongest baseline, while FigEx2 further improves detection in both domains. On PhysSci-Fig-Cap-Test, FigEx2 improves mAP@0.5:0.95 from 0.372 to 0.453 over DAB-DETR; on ChemSci-Fig-Cap-Test, it improves mAP@0.5:0.95 from 0.357 to 0.446 over DAB-DETR. Two factors support this transfer: scientific compound figures share similar spatial layouts across disciplines, so the panel-localization skill carries over; and the attention-level inductive bias from the Entity-Attention KL transfers well, with the detector remaining focused on entity tokens even on out-of-domain captions.

\paragraph{Few-shot prompting.}
We evaluate few-shot prompting by providing ground-truth subcaptions for early panels as exemplars at inference time, using panel A for 1-shot and panels A--B for 2-shot, then evaluating only panels C and later. Table~\ref{tab:fewshot_phy_chem} shows that adding exemplars consistently improves captioning quality in both domains, and the gain is especially pronounced for FigEx2: a few example captions provide the domain vocabulary FigEx2 lacked under zero-shot transfer, lifting BERTScore above the Qwen3-VL-8B baseline at 2-shot (92.22 against 92.06 on PhysSci-Fig-Cap-Test) and indicating that few-shot prompting effectively alleviates the domain-knowledge gap. On ChemSci-Fig-Cap-Test, FigEx2-8B BLEU-4 increases to 46.90 in 2-shot against 42.56 for Qwen3-VL-8B, and 2-shot Entity-F1 reaches 43.81 against 40.95. On PhysSci-Fig-Cap-Test, FigEx2 achieves the best 2-shot Entity-F1 of 35.48 against 31.76. Overall, FigEx2 obtains the strongest entity-level few-shot performance in both domains.

\paragraph{Discussion.}
A consistent pattern holds across settings: FigEx2's lexical scores scale with domain proximity, while its entity-level advantage persists on every benchmark. This is the intended effect of the Entity-Attention KL, which anchors the detector on entity tokens rather than generic descriptive language and thus preserves scientific specificity under domain shift.

\section{Conclusion}
\label{sec:conclusion}
We propose FigEx2, a visual-conditioned framework that jointly detects labeled panels and generates panel-wise captions directly from compound figures, converting otherwise unusable figures into aligned panel-text pairs for downstream pretraining and retrieval. FigEx2 links captioning to detection through a \texttt{[DET]} trigger and feeds the detector hidden states of generated caption tokens, while an Entity-Attention KL aligns the detector's query-to-text attention with entity tokens, providing a stable conditioning signal that improves panel detection in addition to captioning. GRPO with a panel-level Entity-F1 reward then directly optimizes the caption policy for scientific faithfulness. On BioSci-Fig-Cap and our physics/chemistry test suites, FigEx2 improves both detection and captioning over strong fine-tuned baselines, with the largest gains on Entity-F1. Across domain shifts, detection gains hold uniformly thanks to shared spatial layouts and the transferable Entity-Attention KL, while captioning gains scale with domain proximity on lexical metrics but the entity-level advantage persists everywhere. Few-shot prompting further alleviates the remaining domain-knowledge gap, showing that FigEx2 can adapt to new scientific conventions with minimal supervision.

\section*{Ethics Statement}
BioSci-Fig-Cap and our physics and chemistry test suites are derived from open-access sources and do not contain private or sensitive personal data. Curation and caption refinement were performed by trained annotators with domain expertise. While pretrained models may have seen parts of the scientific literature, we mitigate potential leakage by evaluating on held-out splits and reporting results transparently.

\section*{Limitations}
The detection branch relies on DAB-DETR, which predicts axis-aligned bounding boxes. While this suits grid-like layouts, it may be too rigid for non-rectangular regions or heavily overlapping panels common in some scientific domains; segmentation-based delineation is a natural next step. On the data side, our benchmarks cover biomedical, physics, and chemistry literature, but other scientific domains such as materials science, geosciences, and engineering remain uncovered; broadening BioSci-Fig-Cap-style curation to these domains is left to future work. 

\section*{Acknowledgments}
This study was supported by grants from the National Institutes of Health U01CA279618 and R21GM155774 to Y. Huang and in part by the University of Pittsburgh Center for Research Computing, RRID:SCR\_022735. Specifically, this work used the HTC cluster, which is supported by S10OD028483.

\bibliography{custom}

\appendix

\section{Additional Benchmark Details}
\label{app:benchmark_details}

This appendix provides curation and annotation details for the panel-caption benchmarks used in Section~\ref{sec:datasets_cap}.

\subsection{Curation Details}
\label{app:curation}

BioSci-Fig-Cap is curated directly from BioSci-Fig to improve panel-level supervision for labeled, panel-specific captioning. We start from the panel captions and label annotations provided in BioSci-Fig, and apply additional filtering, cleaning, and targeted expert revision to reduce noise and improve terminology consistency and scientific plausibility.

\paragraph{Starting point.}
We inherit the BioSci-Fig panel boxes/labels and their associated panel captions as the initial supervision. We normalize alphabetic labels to uppercase (A--Z) and remove panels without a valid alphabetic label.

\paragraph{Filtering.}
We remove figures where any retained panel caption is extremely short (fewer than five tokens after tokenization), as such lines are typically underspecified. We also remove panel captions that are clearly non-informative, including boilerplate-dominated text and generic references that do not convey panel-specific content.

\paragraph{Cleaning.}
We remove noisy or misaligned panel captions, including mislabeled entries and captions that do not correspond to the assigned panel content. We also remove duplicates and near-duplicates that arise from repeated or redundant panel descriptions.

\paragraph{Expert revision.}
The biology-related panel captions are reviewed by researchers with computational biology and machine learning backgrounds. Revisions focus on improving clarity, terminology consistency, and obvious scientific errors while preserving the intended meaning. We avoid edits that would introduce new claims not supported by the figure.

\paragraph{Cross-domain test set construction.}
PhysSci-Fig-Cap-Test and ChemSci-Fig-Cap-Test are constructed as test-only benchmarks using the FigEx caption decomposition and label association pipeline~\cite{song2025figex}. We select compound figures with high-quality full captions and apply the same FigEx-style procedure to derive label-conditioned panel lines and associate them with alphabetic panel labels.

\subsection{Annotation and Quality Control}
\label{app:annotation}
We use Label Studio for manual review and quality control~\cite{labelstudio}.

\paragraph{Annotation instructions.}
Annotators draw a bounding box around each panel and record its alphabetic label. If a figure contains multiple panels with the same label, annotators reject the compound figure.

\paragraph{Overall accuracy of curated benchmarks.}
We assess curated benchmark quality with a sample-level overall accuracy: for each benchmark, an annotator inspects 200 samples (the full test set for PhysSci-Fig-Cap-Test and ChemSci-Fig-Cap-Test; 200 randomly sampled figures for BioSci-Fig-Cap) and marks each as correct only if its panel boxes and panel captions are judged acceptable. Table~\ref{tab:overall_accuracy} reports the resulting overall accuracy.

\begin{table}[t]
\centering
\small
\setlength{\tabcolsep}{8pt}
\renewcommand{\arraystretch}{1.05}
\begin{tabular}{l c}
\toprule
\textbf{Dataset} & \textbf{Overall Accuracy (\%)} \\
\midrule
BioSci-Fig-Cap        & 97.0 \\
PhysSci-Fig-Cap-Test  & 95.5 \\
ChemSci-Fig-Cap-Test  & 94.5 \\
\bottomrule
\end{tabular}
\caption{Overall accuracy on the curated benchmarks.}
\label{tab:overall_accuracy}
\end{table}

\subsection{Entity Extractors and Matching Policy}
\label{app:entity_extractors}
Each benchmark ships with per-panel entity tokens used by the Entity-Attention KL (Section~\ref{sec:entity_attn_kl}), the Entity-F1 reward (Section~\ref{para:stage4}), and the Entity-F1 evaluation metric (Section~\ref{sec:metrics}). The benchmarks use different domain-specialized NER models. For the biomedical benchmarks (BioSci-Fig-Cap and MedICaT), entities are extracted with BERN2~\cite{sung2022bern2}, and Entity-F1 is computed by matching entities on BERN2's normalized identifiers, which is robust to surface variation (e.g., synonyms, abbreviations) within a single normalized concept. For ChemSci-Fig-Cap-Test we use \texttt{alvaroalon2/biobert\_chemical\_ner}, a community fine-tuned BioBERT~\cite{lee2020biobert} chemical-entity tagger, and for PhysSci-Fig-Cap-Test we use \texttt{pranav-s/PolymerNER}~\cite{shetty2023general}, a MaterialsBERT-based model for polymer and materials entities. Since neither provides a normalized identifier scheme, Entity-F1 on these two benchmarks matches on the surface mention strings produced by the extractor, lower-cased and whitespace-normalized.

\section{Evaluation Protocol Details}\label{app:eval_protocol}

\paragraph{Occurrence-level label alignment.} Ground-truth panels are ordered by the ground-truth reading order. For each label $\ell$, let $\mathcal{R}_\ell=[r_{\ell,1},\ldots,r_{\ell,n_\ell}]$ be reference captions for all occurrences of $\ell$, and let $\mathcal{P}_\ell=[p_{\ell,1},\ldots,p_{\ell,m_\ell}]$ be predicted captions with label $\ell$. We match occurrences by index: $r_{\ell,k}$ pairs with $p_{\ell,k}$ if $k\le m_\ell$, otherwise the prediction for this occurrence is missing; if $m_\ell>n_\ell$, the extra predicted occurrences are treated as extra pairs.

\paragraph{Union pairing and zero scoring.} We form evaluation pairs over the union of ground-truth and predicted labeled lines. A missing pair or an extra pair receives a score of 0 for all caption metrics, including Entity-F1.

\paragraph{Entity-F1 computation.} For each evaluation pair $(\hat{c}, c)$, we run the dataset-specific entity extractor (Appendix~\ref{app:entity_extractors}) on both sides to obtain entity sets $E(\hat{c})$ and $E(c)$, and compute set-level F1 under that dataset's matching policy (ID match for biomedical benchmarks; mention match for chemistry and physics). Empty entity sets on both sides give an F1 of 1; empty on only one side gives 0.

\paragraph{Aggregation.} For each figure, we compute caption metrics on all evaluation pairs and average within the figure as $\mathrm{Score}(\text{figure})=\frac{1}{N}\sum_{i=1}^{N} s(p_i,r_i)$, then report the dataset score as the mean over figures.

\section{Prompts}
\label{app:prompts}

\subsection{Prompt Used for Panel Captioning}
\label{appendix:prompt_caption}
\begin{tcolorbox}[
  enhanced,
  breakable,
  parbox=false,
  colback=gray!15,
  colframe=black,
  width=0.95\columnwidth,
  boxrule=0.8pt,
  halign=left,
  left=4pt, right=4pt, top=4pt, bottom=4pt
]
\textbf{PROMPT\_CAPTIONING}\\
You are given a scientific compound figure.\\
Task: detect subfigures and, for each detected subfigure that shows a visible alphabetic label A to Z or a to z, write exactly one short scientific caption.\\
Formatting rules:\\
1) Output one line per subfigure in ascending label order A, B, C, \ldots\\
2) Use uppercase labels and the exact format: ``A: \textless caption\textgreater''.\\
3) After listing all subfigure captions, output a single [DET] token on a NEW line.\\
Return ONLY the caption lines followed by the final [DET]; no extra text.
\end{tcolorbox}

\subsection{Prompt Used for Panel Detection}
\label{appendix:prompt_det}
\begin{tcolorbox}[
  enhanced,
  breakable,
  parbox=false,
  colback=gray!15,
  colframe=black,
  width=0.95\columnwidth,
  boxrule=0.8pt,
  halign=left,
  left=4pt, right=4pt, top=4pt, bottom=4pt
]
\textbf{PROMPT\_DETECTION}\\
You are given a scientific compound figure containing multiple sub-panels A, B, C, \ldots\\
Detect all sub-panels and output ONLY a JSON array.\\
Each element must be an object with fields:\\
- ``class'': an integer in [0, 25] where A maps to 0, B maps to 1, \ldots, Z maps to 25\\
- ``bbox\_2d'': [x\_min, y\_min, x\_max, y\_max] using normalized coordinates in range [0, 1000]\\
Rules:\\
- Do not output any text outside the JSON array.\\
- bbox\_2d must satisfy x\_min \textless x\_max and y\_min \textless y\_max.\\
- Include every detected panel; multiple boxes may share the same class.
\end{tcolorbox}

\section{Additional Implementation Details}
\label{app:impl}

We use LoRA with rank 16, alpha 32, dropout 0.05, and target modules
\texttt{q\_proj, k\_proj, v\_proj, o\_proj, gate\_proj, up\_proj, down\_proj}
as a parameter-efficient adaptation method.
We train with batch size 1 and gradient accumulation 16, using a maximum
sequence length of 8192 tokens throughout training and inference.

We report BERTScore as $100\times$F1.

\section{Examples}

\begin{figure*}[b]
    \centering
    \includegraphics[width=0.9\textwidth]{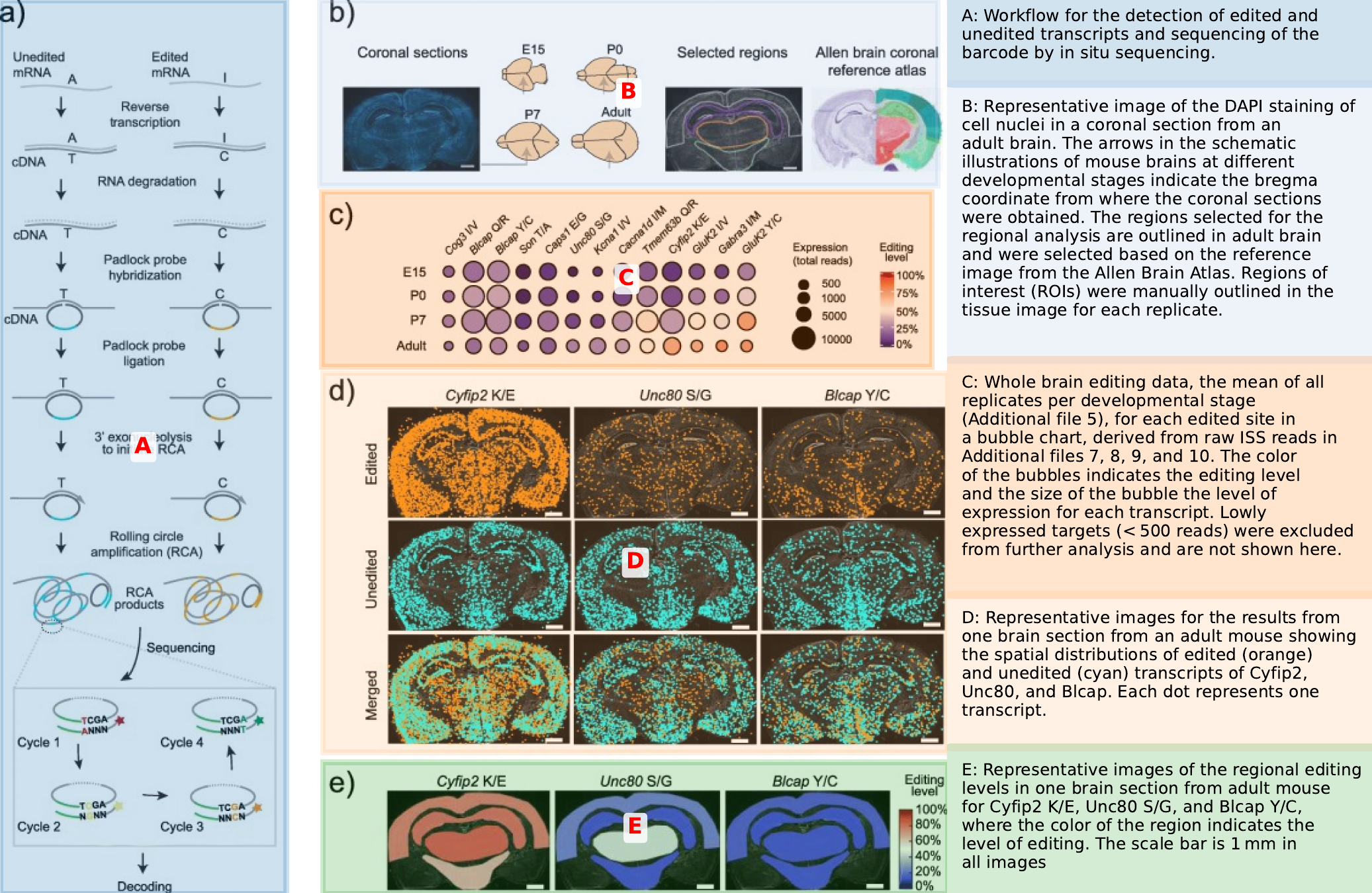}
    \caption{Compound figure with panel boxes with aligned subcaptions (example 1).}
    \label{fig:example1}
\end{figure*}

\begin{figure*}
    \centering
    \includegraphics[width=0.9\textwidth]{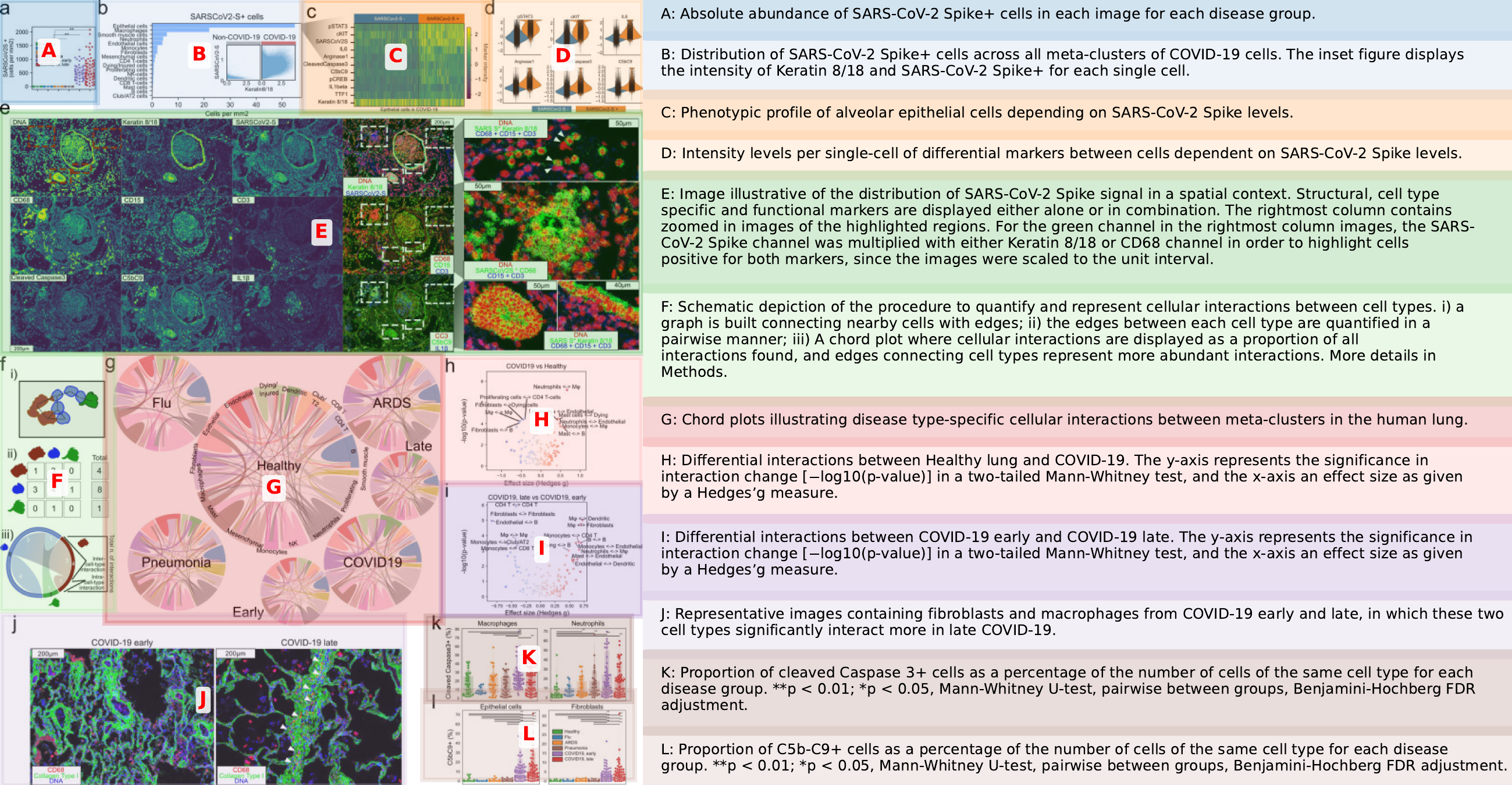}
    \caption{Compound figure with panel boxes with aligned subcaptions (example 2).}
    \label{fig:example2}
\end{figure*}

\begin{figure*}
    \centering
    \includegraphics[width=0.9\textwidth]{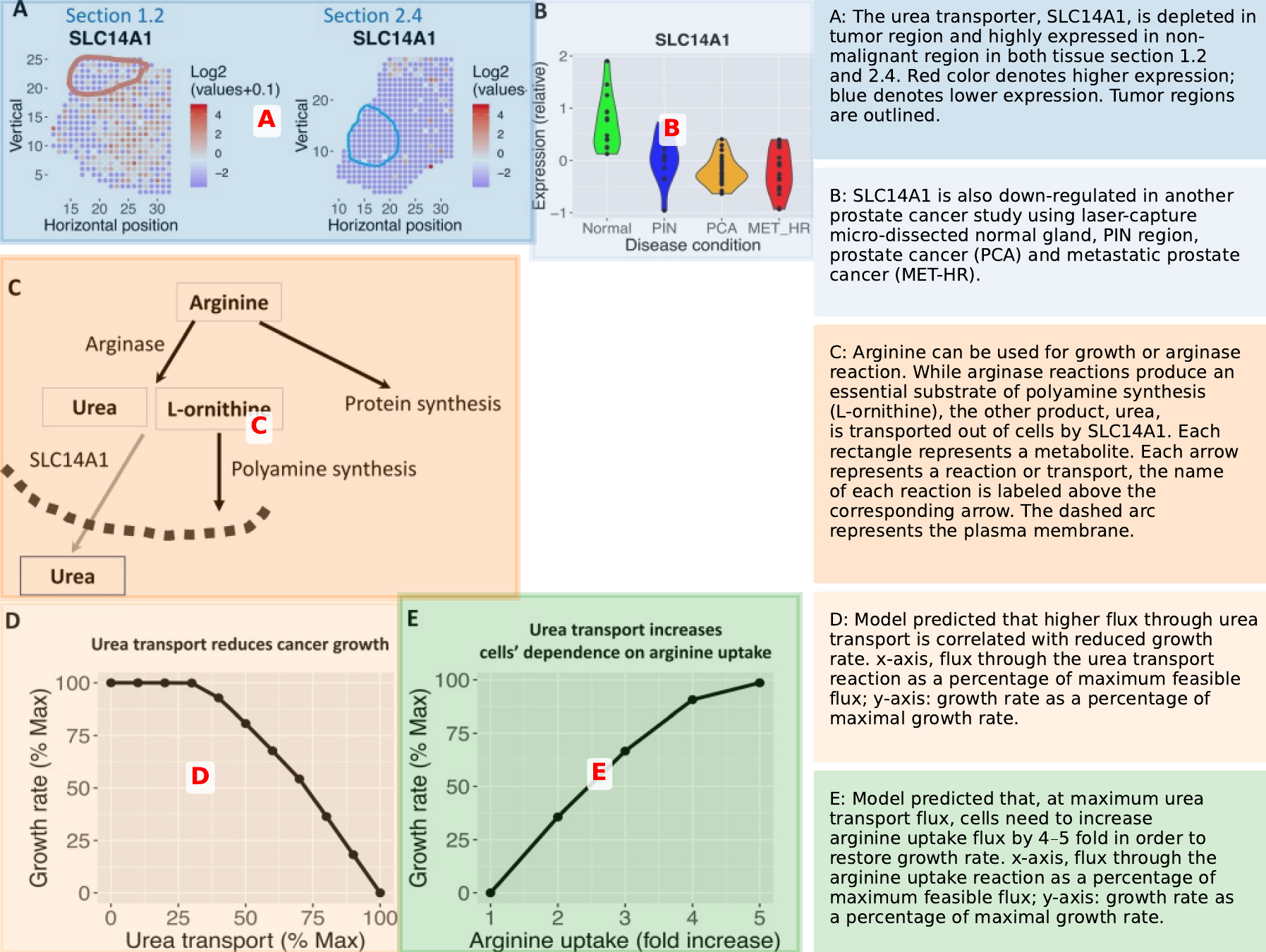}
    \caption{Compound figure with panel boxes with aligned subcaptions (example 3).}
    \label{fig:example3}
\end{figure*}

Figures~\ref{fig:example1}--\ref{fig:example3} show qualitative examples with the compound image and the panel boxes and captions.

\end{document}